%% file: main.tex
\documentclass[10pt,twocolumn,letterpaper]{article}

\usepackage{cvpr}              

\usepackage{amsmath} 
\usepackage{lineno}
\usepackage{bm}
\usepackage[nice]{nicefrac}
\usepackage{amssymb}
\usepackage{newfloat}
\usepackage{listings}
\usepackage{colortbl}
\usepackage{multirow}
\usepackage{makecell}
\usepackage{dsfont}
\usepackage{arydshln} 
\usepackage{overpic}
\usepackage{xcolor}

\usepackage[margin=1in]{geometry}
\usepackage{graphicx}
\usepackage{caption}
\usepackage{subcaption}
\usepackage{multirow}
\usepackage{array}
\usepackage{rotating}
\usepackage{booktabs}
\usepackage{adjustbox}
\usepackage{tikz}
\usepackage{dashrule}
\usepackage[accsupp]{axessibility}  
\usepackage{pifont}

\usepackage{cuted} 
\usepackage{wrapfig}
\definecolor{cvprblue}{rgb}{0.21,0.49,0.74}
\definecolor{color3}{RGB}{255, 255, 200}
\definecolor{color2}{RGB}{255, 230, 180}
\definecolor{color1}{RGB}{255, 181, 163}

\usepackage{amsmath,amssymb}
\usepackage{algorithm}
\usepackage{algpseudocode}
\input{preamble}

\definecolor{cvprblue}{rgb}{0.21,0.49,0.74}
\usepackage[pagebackref,breaklinks,colorlinks,allcolors=cvprblue]{hyperref}

\def\confName{CVPR}
\def\confYear{2026}

\title{SGS-Intrinsic: Semantic-Invariant Gaussian Splatting for Sparse-View Indoor Inverse Rendering}

\author{
    Jiahao Niu$^1$ \quad Rongjia Zheng$^1$ \quad Wenju Xu$^2$ \quad Wei-Shi Zheng$^{1,3}$ \quad Qing Zhang$^{1,3}$\thanks{Corresponding author.} \\
    \small $^1$ Sun Yat-sen University, China \quad $^2$ Amazon, USA \\
    \small $^3$ Key Laboratory of Machine Intelligence and Advanced Computing, Ministry of Education, China
}

\begin{document}

\twocolumn[
    \maketitle
    \vspace{-3.0em}
    \input{images/teaser}
    \bigbreak
]

\begingroup
\renewcommand\thefootnote{}
\footnotetext{*Corresponding author (zhangq93@mail.sysu.edu.cn).}
\endgroup

\input{sec/0_abstract}    
\input{sec/1_intro}

\input{sec/2_related_work}

\input{sec/3_method}

\input{sec/4_experiment}

\input{sec/5_conclusion}

\vspace{0.5em}
\noindent\textbf{Acknowledgement.} This work was partially supported by the National Natural Science Foundation of China (62471499), the Guangdong Basic and Applied Basic Research Foundation (2023A1515030002).

{
    \small
    \bibliographystyle{ieeenat_fullname}
    \bibliography{main}
}
\input{sec/X_supple}


\end{document}

%% file: preamble.tex
\usepackage{multirow}



%% file: images/teaser.tex
\begin{center}
    \includegraphics[width=1\textwidth]{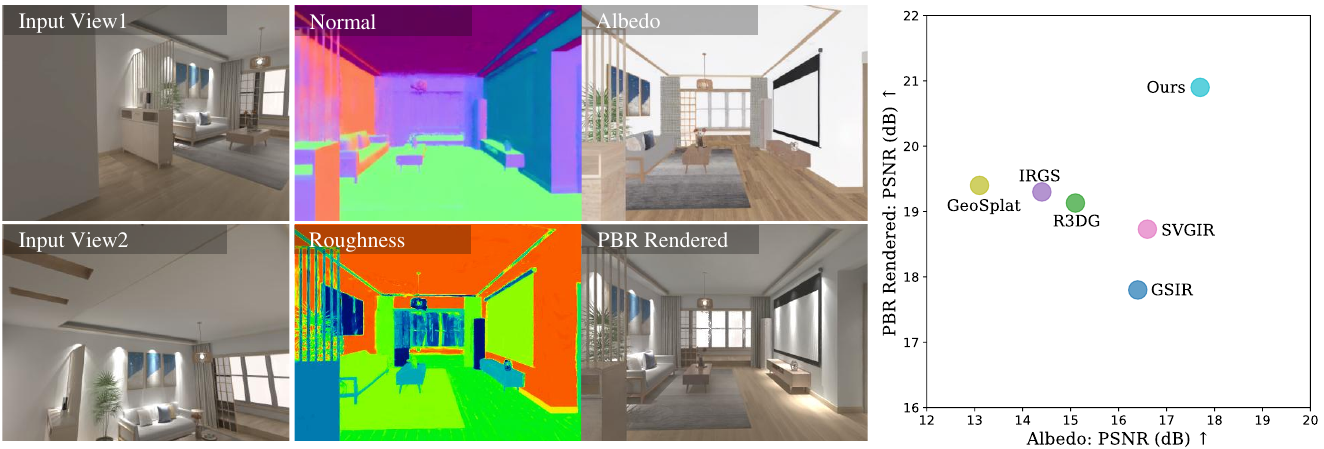} \\
    \vspace{-2mm}
    \captionof{figure}{
        \textbf{Sparse-view indoor inverse rendering.} Our method achieves high-quality scene-level disentanglement of illumination and material properties from sparse-view input images. As can be seen, our method allows to produce high-quality novel-view PBR rendering, and also outperforms previous methods, e.g., GeoSplat~\cite{ye2025geosplatting}, IRGS~\cite{gu2025irgs}, R3DG~\cite{gao2024relightable}, SVGIR~\cite{sun2025svg}, and GSIR~\cite{liang2024gs} on the Interiorverse dataset~\cite{zhu2022learning}.
    }
    \label{fig:teaser}
\end{center}

%% file: sec/0_abstract.tex
\begin{abstract}

We present SGS-Intrinsic, an indoor inverse rendering framework that works well for sparse-view images. Unlike existing 3D Gaussian Splatting (3DGS) based methods that focus on object-centric reconstruction and fail to work under sparse view settings, our method allows to achieve high-quality geometry reconstruction and accurate disentanglement of material and illumination. The core idea is to construct a dense and geometry-consistent Gaussian semantic field guided by semantic and geometric priors, providing a reliable foundation for subsequent inverse rendering. Building upon this, we perform material–illumination disentanglement by combining a hybrid illumination model and material prior to effectively capture illumination–material interactions. To mitigate the impact of cast shadows and enhance the robustness of material recovery, we introduce illumination-invariant material constraint together with a deshadowing model. Extensive experiments on benchmark datasets show that our method consistently improves both reconstruction fidelity and inverse rendering quality over existing 3DGS-based inverse rendering approaches. Our code is available at \url{https://github.com/GrumpySloths/SGS_Intrinsic.github.io}.

\end{abstract}

%% file: sec/1_intro.tex
\section{Introduction}
\label{sec:intro}

Multi-view inverse rendering aims to decompose material, geometry, and lighting from a set of input-view images (see Figure~\ref{fig:teaser}), which is of great significance for understanding the entire 3D scene and can be extensively applied to various downstream tasks such as material editing, relighting, AR, VR, and more~\cite{liang2024gs,lai2025glossygs,wang2023prolificdreamer,chung2023luciddreamer,zheng2025dnf}. However, inverse rendering under sparse views, especially for indoor scenes, remains a highly ill-posed problem due to the sparsity of supervision signals~\cite{liao2025spc,wu2025sparse2dgs}, scene complexity, and the strong coupling of material, lighting, and geometry~\cite{du2024gs,liang2024gs}.

Specifically, sparse-view indoor inverse rendering using Gaussian Splatting that we target in this paper, faces the following three challenges: (i) Under sparse-view settings, Gaussian reconstruction struggles to obtain reliable geometry, adversely affecting the effectiveness of subsequent inverse rendering. (ii) previous utilized illumination models are usually ineffective for characterizing complex indoor lighting conditions, and the lack of sufficient material priors makes it difficult to disentangle material and lighting, (iii) inverse rendering from sparse views poses challenge to accurately model specular highlights and cast shadows, making shadows  often erroneously baked into material.

Numerous approaches~\cite{wang2024use,wu2025sparse2dgs,zhu2024fsgs,fan2024instantsplat,bao2025free360,wang2024freesplat,wu2024reconfusion,wu2025difix3d+,liu20243dgs} have explored Gaussian-based geometry reconstruction under sparse-view settings. However, most of them focus solely on general geometric reconstruction without considering material disentanglement. Moreover, they primarily target the recovery of high-frequency geometric fields, struggle to simultaneously recover material and geometric \cite{liao2025spc}. 

Another challenge for scene-level inverse rendering lies in illumination modeling. Prior works~\cite{liang2024gs,chen2024gi,sun2025svg,gu2025irgs,ye2025geosplatting,gao2024relightable,lai2025glossygs,du2024gs} typically assume distant illumination sources --- an assumption that provides a reasonable approximation at the object level but fails to capture near-field, high-frequency lighting effects that are essential for accurate reconstruction of indoor scenes.

In this paper, we present SGS-Intrinsic, a two-stage framework for sparse-view indoor inverse rendering. Unlike traditional SfM-based approaches~\cite{schonberger2016structure}, which yield only sparse point cloud that hinder reliable Gaussian optimization~\cite{zhu2024fsgs,wu2025sparse2dgs}, we first leverage VGGT~\cite{wang2025vggt} to construct a dense scene layout of point clouds. To further compensate for the limited supervision, we incorporate normal and semantic priors distilled from a pretrained model into 3D Gaussian representation. Besides, a semantic consistency constraint between virtual and training views is introduced to alleviate overfitting. Next, we perform inverse rendering with a hybrid lighting model~\cite{du2024gs}, combining environment maps for ambient illumination and SGMs~\cite{wang2009all} for high-frequency lighting. To achieve effective material–illumination disentanglement, we leverage diffusion-based material priors and enforce cross-view and cross-illumination consistencies, harvesting lighting- and view-invariant material reconstruction. Finally, we employ a lightweight deshadowing module to explicitly model visibility, preventing shadows from being baked into albedo.

In summary, our contributions are as follows:
\begin{itemize}[leftmargin=2em]
\setlength\itemsep{0.5em} 
\item 
We present an inverse rendering framework that enables high-quality geometry reconstruction and accurate disentanglement of material and illumination from sparse-view indoor images. 
\item
We introduce several additional supervision signals to enable reliable inverse rendering under sparse views setting, and propose to resolve the issue of cast shadows being baked into material through an effective and lightweight deshadowing model.
\item
Extensive experiments show that our method significantly outperforms state-of-the-art approaches on sparse-view indoor inverse rendering.
\end{itemize}

%% file: sec/2_related_work.tex
\section{Related Work}

\begin{figure*}[!t]
\centering
\includegraphics[width=1\textwidth]{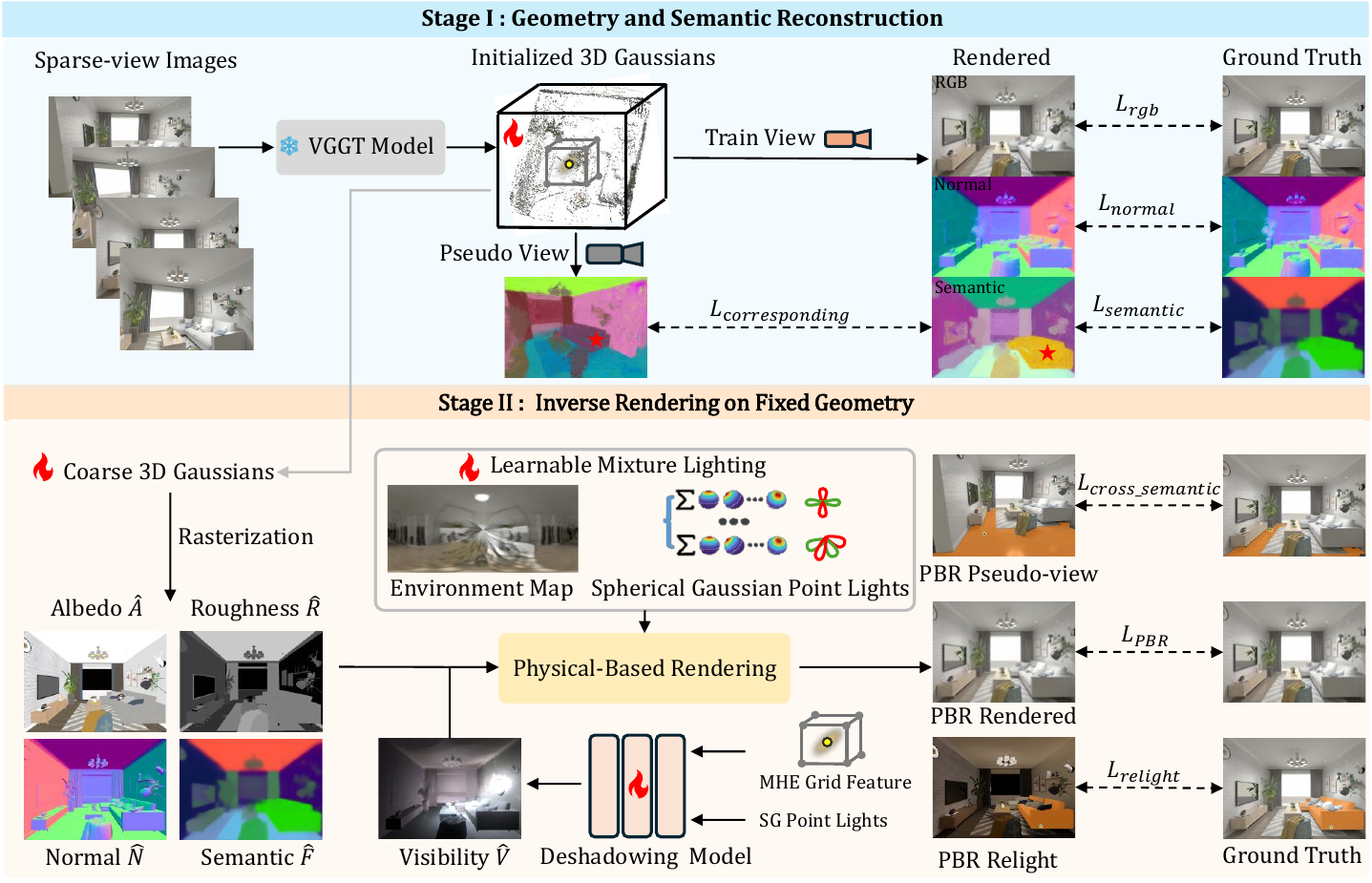} \\
\vspace{-3mm}
\caption{\textbf{Overview of our method.}
There are two-stage training in our framework. In stage I, we leverage a pretrained VGGT model to get a dense scene layout point cloud. Next, the geometry of the 3D Gaussians is supervised by normal and semantic priors distilled from a pretrained model. In state II, we perform inverse rendering based on the Gaussians obtained from the first stage. A mixture light model together with a deshadowing module are employed to model illumination and occlusion relationships, which are then integrated with the Gaussian-rendered G-buffer for physically based (PBR) novel view rendering. Moreover, to ensure consistent material representations during training, we incorporate an additional material consistency constraint.}
\label{fig:pipeline}
\end{figure*}

\textbf{Geometry reconstruction from sparse views.} 
 Existing sparse-view geometry reconstruction methods can be broadly categorized into two groups: per-scene optimization methods and pre-training based methods. Per-scene optimization methods ~\cite{li2024dngaussian,zhu2024fsgs,wang2024use,pooleDreamFusionTextto3DUsing2022,chung2024depth,zhang2024cor,paul2024sp2360,peng2024structure,han2024binocular} focus on learning to recover geometry of a specific scene, and thus cannot generalize to other scenes. In contrast, by taking full advantage of powerful priors from pre-trained models, pretraining based methods allow to generalize beyond the training scenes \cite{wang2024dust3r,wang2025vggt,leroy2024grounding}. Despite the effectiveness of pretraining based methods, they basically lack the ability to achieve material-illumination disentanglement.

\vspace{0.5em}
\noindent \textbf{Inverse Rendering.}
Inverse rendering has long been a fundamental problem in computer vision and computer graphics.
Early methods focus on predicting intrinsic properties from a single image. Representative works include ~\cite{10.1145/3641519.3657445,careaga2024colorful,liang2025diffusion,chen2024intrinsicanything,zeng2024dilightnet,careaga2023intrinsic,kocsis2024intrinsic}. However, intrinsic decomposition from a single image inherently lacks multi-view consistency and often struggles with out-of-distribution cases \cite{du2024gs}.
Hence, recent methods \cite{du2024gs,liang2024gs,gu2025irgs,shi2025gir,wynn2023diffusionerf,ni2025decompositional} aim to recover a 3D-consistent intrinsic field from multiple input images. Although these methods demonstrate impressive results, they focus primarily on object-level scenes, and fail to work well under sparse view conditions involving complex indirect lighting and occlusion.

%% file: sec/3_method.tex
\section{Method}

Given sparse-view indoor images as input, our goal is to recover the scene geometry and intrinsic properties based on 3D Gaussian. As shown in Figure~\ref{fig:pipeline}, our method consists of two stages. In stage I, we leverage semantic and geometric priors from pre-trained models to help construct a high-quality Gaussian-based geometric field, while in stage II, we optimize the intrinsic properties with a hybrid illumination model together with a material-consistency regularized training strategy.

\subsection{Geometry and Semantic Reconstruction}
\label{sec:geo_sem_init}

Most previous 3D Gaussian reconstruction methods rely on an RGB reconstruction loss $\mathcal{L}_{rgb}$~\cite{kerbl20233d} for training. However, under sparse-view supervision, they tend to overfit to the observed viewpoints, leading to unreliable geometry reconstruction. To alleviate this issue, we choose to leverage powerful pre-trained models to provide additional supervisory signals. Specifically, we employ StableNormal~\cite{ye2024stablenormal} and LSEG~\cite{li2022language} to provide normal guidance and semantic supervision, respectively. For normal, we define the following loss:
\begin{equation}
    \mathcal{L}_{normal} = 1 - \hat{n}^T n_m,
\end{equation}
where $\hat{n}$ denotes the rendered normal from the Gaussian field, $n_m$ denotes normal predicted by StableNormal.

For semantic supervision, we first encode the input images as pixel-level semantic features $F \in \mathbb{R}^{H \times W \times 512}$ using LSEG encoder. We then employ Qwen3VL~\cite{bai2025qwen3} to identify all object categories $M$, and obtain their corresponding textual embeddings $T \in \mathbb{R}^{M \times 512}$ via CLIP encoder~\cite{radford2021learning}. Next, we compute a semantic label map $S \in \mathbb{R}^{H \times W}$ by $S_{i,j} = \arg\max_{m \in \{1, \dots, M\}} \text{cos}(F_{i,j}, T_m)$, where $\text{cos}(\cdot, \cdot)$ denotes cosine similarity. With semantic features $F$ and the label map $S$, we define the semantic constraint as:
\begin{equation}
    \mathcal{L}_{semantic} = \mathcal{L}_{\text{cos}}(\hat{F}, F) + \mathcal{L}_{\text{ce}}(\hat{S}, S),
\end{equation}

where $\mathcal{L}_{cos}$ and $\mathcal{L}_{ce}$ respectively denote cosine and cross-entropy losses. Following previous work~\cite{liao2025spc}, the rendered semantic feature $\hat{F}$ is obtained via alpha blending, while the rendered segmentation logits are defined as $\hat{S} = \cos ( \hat{F}, T )$.


To avoid overfitting, we introduce a semantic consistency constraint between training and pseudo (virtual) views, which is formulated as:
\begin{align}
    \mathcal{L}_{\text{corresponding}} = 
    \sum_{p=1}^{P} 
    \left\| 
    \frac{\hat{F}_p \odot \mathcal{M}_p^*}{\sum \mathcal{M}_p^*} -
    \frac{\hat{F} \odot \mathcal{M}}{\sum \mathcal{M}} 
    \right\|_2    \notag  \\
    -  
    \sum_{p=1}^{P} 
    \mathbb{R}[\Phi(\hat{S} \odot \mathcal{M})]
    \log(\hat{S}_p^* \odot \mathcal{M}_p^*),
\end{align}
where $P$ denotes the number of sampled pseudo-views (see the supplementary for details). $\mathcal{M}$ and $\mathcal{M}_p^*$ represent SAM~\cite{ravi2024sam} generated masks for the training and pseudo-views, respectively. $\odot$ denotes the element-wise product. Here, $\hat{F}$ and $\hat{S}$ refer to the rendered feature map and semantic label map of the training view, while $\hat{F}_p$ and $\hat{S}_p^*$ denote their corresponding counterparts of the $p$-th pseudo-view. $\Phi$ is a semantic uniformity operation defined as $\Phi(\hat{S} \odot \mathcal{M})
= \arg\max_{c} \sum_{i \in \mathcal{M}} \mathbf{1}\!\left\{ \arg\max\bigl(\hat{S}_i)\bigr) = c \right\}$, where $c$ is the dominant class within the region mask $\mathcal{M}$, while $\mathbb{R}[\cdot]$ reshapes tensors to match the mask dimension. The goal of this operation is to apply a majority voting mechanism over each region to enforce semantic consistency.

The overall loss function for stage I is expressed as:
\begin{equation}
    \mathcal{L}_{S1} = \mathcal{L}_{rgb}+\mathcal{L}_{semantic} + \lambda_1 \mathcal{L}_{normal} + \lambda_2 \mathcal{L}_{corresponding},
\end{equation}
where we set $\lambda_1=0.05$, $\lambda_2=0.3$ in our experiments.

\input{tables/evaluation_interiorverse}

\subsection{Inverse Rendering on Fixed Geometry}

Previous methods often suffer from degraded performance at the scene level due to limited expressiveness of the illumination model and unrealiable geometry reconstruction under sparse-view settings. To address these issues, we introduce a hybrid illumination model and a lightweight deshadowing module, which enables explicitly disentangling and modeling illumination as well as occlusion, respectively. Furthermore, to facilitate the decoupling of illumination and material, we introduce a material-consistent regularized training strategy.

\vspace{0.5em}
\noindent\textbf{Hybrid illumination model.}
To model complex lighting effects, we propose to cast the incident light $L_i$ as a combination of learnable point lights $L_i^{\text{SGM}}$ and differentiable environment illumination $L_i^{\text{env}}$, representing the high and low frequency lighting components:
\begin{equation}
L_i = L_i^{\text{SGM}} + L_i^{\text{env}},
\end{equation}
where the point-light component $L_i^{\text{SGM}}$ is represented by scattered Spherical Gaussians (SGMs)~\cite{du2024gs}:
\begin{equation}
\small
\begin{aligned}
\text{SGM}(\omega_i; b, \lambda, \mu) =
\sum_{k=1}^{N_{\text{sg}}} \mu_k e^{\lambda_k (\omega_i \cdot b_k^{-1})} \cdot W_k,
\end{aligned}
\end{equation}
where $b_k$, $\mu_k$, and $\lambda_k$ denote the axis, amplitude, and sharpness of the $k$-th Spherical Gaussian (SG), respectively. $W_k$ represents the blending weight of each SG component within the light source, $N_{\text{sg}}$ is the total number of SG lobes employed in the mixture. For the environment illumination component $L_i^{\text{env}}$, following~\cite{liang2024gs}, we represent it as a learnable cubemap with a spatial resolution of $6 \times 512 \times 512 \times 3$.

\input{images/visibility_analysis}

\vspace{0.5em}
\noindent \textbf{Lightweight deshadowing module.}

As illustrated in Figure~\ref{fig:visibility_analysis}, the inherent difficulty of recovering accurate geometry from sparse-view images makes it difficult to obtain reliable occlusion relationships, which subsequently hinders the faithful modeling of indirect illumination. Hence, we introduce a lightweight occlusion field to predict the visibility between surface points and point lights explicitly. Specifically, we employ a multi-level hash encoder $\varepsilon_o$ ~\cite{muller2022instant} to obtain multi-resolution feature encoding of the 3D scene, where a spatial position $\mu_i$ is mapped to an occlusion feature $e_i=\varepsilon_o(\mu_i)$. By encoding the entire 3D space at multiple resolutions and performing interpolated sampling for the given spatial points, we alleviate the issue of inaccurate Gaussian-based geometric modeling and occlusion estimation that commonly arise under sparse-view settings. The encoded multi-scale spatial features $e_i$, together with the relative direction $d_{ji} = \frac{p_j -\mu_i}{\|p_j - \mu_i\|_2}$ and distance $\delta_{ji} = \|p_j-\mu_i\|_2$ of a point light $j$, are then fed into a lightweight decoder $f(\cdot)$ to predict the visibility by $V_{ji} = f(e_i,\delta_{ji},d_{ji})$. This design enables the model to perceive the occlusion relationships between spatial points and local light sources, allowing it to infer the general direction of cast shadows and disentangle them from intrinsic material appearance.

\vspace{0.5em}
\noindent\textbf{Material-consistent regularized training.}
Inverse rendering is inherently ill-posed due to the strong coupling between illumination and material properties. Simply leveraging material priors from pretrained models for predicting the material attributes of Gaussian primitives is still insufficient, as the learned material (albedo) attributes often fall into undesired local optima (see Figure~\ref{fig:ablation_albedo_estimation}). To address this issue, we propose to construct additional constraints based on the illumination invariance and multi-view consistency of material properties.

Specifically, we introduce illumination perturbations under a fixed viewpoint to enforce albedo invariance. Given multiple lighting conditions, the rendered albedo $\hat{A}$ by alpha blending~\cite{liang2024gs} should be unchanged. Following~\cite{careaga2024colorful}, we define the relighting consistency loss as:
\begin{equation}
    \small
\begin{aligned}
&\mathcal{L}_{relight} =  \sum_{i=1}^{N} 
\left(
\mathcal{L}_{mse}\!\left(\hat{A}, A_i, \mathcal{M}_{o}\right)
+ \mathcal{L}_{msg}\!\left(\hat{A}, A_i, \mathcal{M}_{o}\right)
\right), \\
&\mathcal{L}_{mse}(\hat{A}, A_i, \mathcal{M}_{o}) 
= \left\| (\hat{A} - A_i) \odot \mathcal{M}_{o} \right\|_2, \\
&\mathcal{L}_{msg}(\hat{A}, A_i, \mathcal{M}_{o})
= \sum_{l=1}^{L} 
\left\| (\nabla \hat{A}^{l} - \nabla A_i^{l}) \odot \mathcal{M}_{o} \right\|_2,
\end{aligned}
\end{equation}
where $A_i$ represents the material prior predicted by the pretrained RGBX model~\cite{10.1145/3641519.3657445} under $N$ sampled illumination perturbations. Detailed sampling strategies for the perturbations are provided in the supplementary. To achieve precise elimination of specular highlights and cast shadows, we obtain an object-level mask $\mathcal{M}_o$ derived from the SAM model, with which we are able to achieve fine-grained, localized supervision, ensuring that the relighting constraints are applied specifically to the target object. The relighting objective consists of two components: a standard MSE loss $\mathcal{L}_{mse}$ and a multi-scale gradient loss $\mathcal{L}_{msg}$. $\nabla$ denotes the spatial gradient operator, while $l$ denotes the specific level within a multi-scale image pyramid.

\input{images/interiorverse_compare}

Considering that the material priors from pretrained models often lack multi-view consistency, we devise a multi-view consistency constraint to regularize the disentanglement of illumination and material attributes, based on the observation that regions sharing similar semantics often have similar material properties.

For each training view, we sample $P$ additional viewpoints using the same pseudo-view strategy adopted in stage~I and employ SAM to extract masks $\mathcal{M}$ and $\mathcal{M}_p^*$ for training and pseudo-views. Next, we formulate the following multi-view consistency loss:
\begin{equation}
\mathcal{L}_{cross\_semantic} = \sum_{p=1}^{P} \left( \left\| \bar{F}_p^* - \bar{F} \right\|_2 + \left\| \bar{A}_p^* - \bar{A} \right\|_2 \right),
\end{equation}
where $\bar{F}$ and $\bar{A}$ denote the mean semantic feature and mean rendered albedo derived from mask $M$ of the training view, defined as $\bar{F} = \frac{(\hat{F} \odot \mathcal{M})}{\sum \mathcal{M}}$ and $\bar{A} = \frac{(\hat{A} \odot \mathcal{M})}{\sum \mathcal{M}}$, respectively. Similarly, $\bar{F}_p^*$ and $\bar{A}_p^*$ represent the corresponding mean semantic feature and albedo rendered of the $p$-th pseudo view, respectively.

Finally, we adopt a photometric constraint $\mathcal{L}_{\text{PBR}}$ from \cite{du2024gs} to enforce the physically-based rendering results to be close to the corresponding ground truths.

The overall loss function for stage II is now defined as:
\begin{equation}
    \mathcal{L}_{S2} = \mathcal{L}_{\text{PBR}} + 
    \lambda_3 \mathcal{L}_{cross\_semantic} +
    \lambda_4 \mathcal{L}_{relight},
\end{equation}
where we empirically set $\lambda_3=0.4$ and $\lambda_4=0.4$.

\subsection{Implementation Details}
We conduct all experiments on a single NVIDIA RTX 4090 GPU. For stage I, we set the total number of training iterations as 7000. Virtual view sampling and the semantic consistency constraint between virtual and real views are activated after 2000 iterations. For stage II, the number of training iterations is set to 3000. We employ CLIP ViT-B/16~\cite{radford2021learning} for semantic feature extraction and SAM2 Hiera-L~\cite{ravi2024sam} for region mask generation. The entire training process consists of 10,000 iterations and takes approximately 40 minutes. For more details, please see the supplementary material.

%% file: tables/evaluation_interiorverse.tex
\begin{table*}[!t]
\centering
\caption{
\textbf{Quantitative evaluation on the Interiorverse dataset.} 
}
\vspace{-0.2cm}
\resizebox{\textwidth}{!}{
    \begin{tabular}{@{}clcc ccc ccc ccc c c}
    \toprule
    & \multirow[b]{2}{*}{Method} &
    \multirow{2}{*}[-6pt]{\makecell[t]{Roughness\\MSE $\downarrow$}} &
    \multirow{2}{*}[-6pt]{\makecell[t]{Normal\\MAE $\downarrow$}} &
    \multicolumn{3}{c}{Albedo} &
    \multicolumn{3}{c}{Novel View Synthesis} &
    \multicolumn{3}{c}{Novel View Synthesis for PBR}&\multirow[b]{2}{*}{FPS} \\
    \cmidrule(lr){5-7}\cmidrule(lr){8-10}\cmidrule(lr){11-13}
    & & & & 
    PSNR $\uparrow$ & SSIM $\uparrow$ & LPIPS $\downarrow$ & 
    PSNR $\uparrow$ & SSIM $\uparrow$ & LPIPS $\downarrow$ & 
    PSNR $\uparrow$ & SSIM $\uparrow$ & LPIPS $\downarrow$  \\
    \midrule
    & GSIR~\cite{liang2024gs} & 28.1 & 26.2 & 16.4 & 0.56 & 0.41 & 18.5 & 0.62 & 0.35 & 17.8 & 0.60 & 0.43&\textbf{40.01} \\
    & GIGS~\cite{chen2024gi} & 30.2 & 25.9 & 12.1 & 0.49 & 0.42 & 18.6 & 0.60 & 0.59 & 17.2 & 0.58 & 0.44&25.26 \\
    & R3DG~\cite{gao2024relightable} & 25.6 & 24.3 & 15.1 & 0.68 & 0.36 & 19.8 & 0.75 & 0.30 & 19.13 & 0.68 & 0.43&9.5 \\
    & IRGS~\cite{gu2025irgs}  & 26.2 & 23.2 & 14.4 & 0.61 & 0.35 & 18.9 & 0.68 & 0.34 & 19.3 & 0.71 & 0.32 &0.20\\
    & SVGIR~\cite{sun2025svg}  & 22.3 & 20.1 & 16.6 & 0.70 & 0.33 & 19.0 & 0.76 & 0.29 & 18.73 & 0.74 & 0.31  &11.83\\
    & GeoSplat~\cite{ye2025geosplatting}  & 17.3 & 18.4 & 13.1 & 0.52 & 0.41 & 20.6 & 0.71 & 0.44 & 19.40 & 0.67 & 0.41& 10.1 \\
    & Ours & \textbf{16.1} & \textbf{17.0} & \textbf{17.7} & \textbf{0.76} & \textbf{0.29} & \textbf{21.7} & \textbf{0.76} & \textbf{0.43} & \textbf{20.90} & \textbf{0.73} & \textbf{0.31} &15.05 \\

    \bottomrule
    \end{tabular}
}

\label{tab:interiorverse_comparison}
\end{table*}

%% file: images/visibility_analysis.tex
\begin{figure}[!t]
\setlength\tabcolsep{1pt}
\centering
\scriptsize 
\includegraphics[width=1\linewidth]{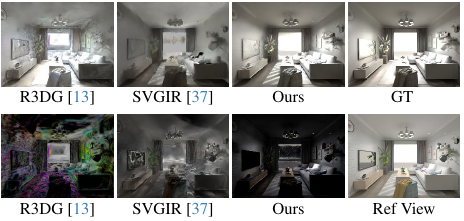} \\
\vspace{-2mm}
\caption{\footnotesize
\textbf{Comparison of scene radiance and local-light rendering for novel views.} As shown, both R3DG and SVGIR fail to accurately model local illumination and occlusion for sparse-view indoor scenes, while our method produces physically plausible rendering with smooth local light effects.
}
\vspace{-3mm}
\label{fig:visibility_analysis}
\end{figure}

%% file: images/interiorverse_compare.tex
\newcommand{\softdashedline}{
  \vspace{-7mm}
  \begin{center}
    \begin{tikzpicture}
      \draw[dashed, line width=0.8pt, color=gray!50] (0,0) -- (\linewidth+0,0);
    \end{tikzpicture}
  \end{center}
  \vspace{-3mm}
}

\begin{figure*}[t]
\setlength\tabcolsep{1.25pt}
\centering
\includegraphics[width=\textwidth]{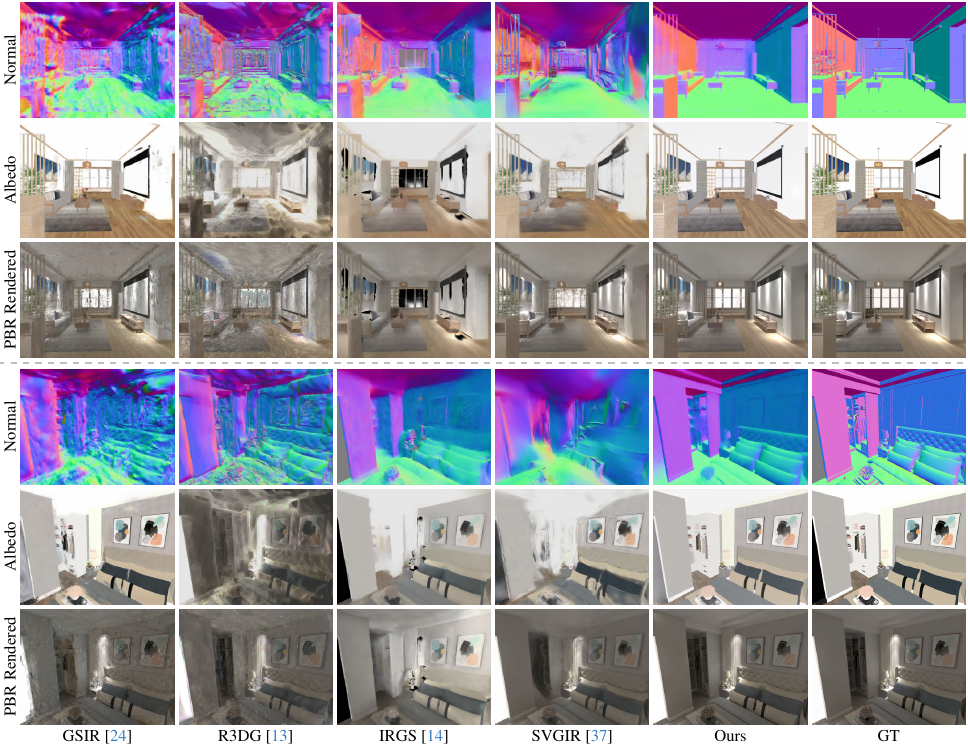}\\
\vspace{-2mm}
\caption{\textbf{Qualitative comparison of inverse rendering on the Interiorverse dataset~\cite{zhu2022learning}.} }
\label{fig:interiorverse_compare}
\end{figure*}

%% file: sec/4_experiment.tex
\section{Experiments}

\input{images/mipnerf_comparison}

\noindent \textbf{Datasets and metrics.}
We evaluate our method on two synthetic datasets including the scene-level Interiorverse dataset~\cite{zhu2022learning} and the object-level TensoIR~\cite{jin2023tensoir} dataset, as well as three real-world datasets including the MipNeRF dataset~\cite{barron2022mip}, the FIPT~\cite{wu2023factorized}, and the DL3DV~\cite{ling2024dl3dv} dataset involving complex lighting conditions. For synthetic datasets, we assess both geometric reconstruction and inverse rendering quality using standard metrics including PSNR, SSIM, and LPIPS~\cite{zhang2018unreasonable}. Note, we provide evaluation on the TensoIR dataset in the supplementary material. For real-world datasets without ground-truth inverse rendering results, we instead perform comparison on novel view synthesis and also employ the MAE~\cite{bae2024rethinking} metric to evaluate normal reconstruction accuracy, MSE~\cite{bai2025qwen3} for roughness. For sparse-view selection, we follow the standard protocol utilized by previous methods~\cite{liao2025spc,fan2024instantsplat,zhu2024fsgs}. For the Interiorverse and TensoIR datasets, we sample 12 and 8 input views for training, while for the MipNeRF, DL3DV and FIPT datasets, we use 24 training views.

\input{images/ablation_analysis}

\subsection{Comparison with State-of-the-Art Methods}

\noindent\textbf{Evaluation on the Interiorverse dataset.}  
Table~\ref{tab:interiorverse_comparison} and Figure~\ref{fig:interiorverse_compare} present comparison between our method and previous state-of-the-art approaches on the synthetic Interiorverse indoor scene dataset. As can be seen, both the quantitative and qualitative results demonstrate that our method not only outperforms existing methods in geometry and material reconstruction, but also produces better PBR rendering results under novel views.

\vspace{0.5em}
\noindent\textbf{Evaluation on the MipNeRF dataset.}
To further validate the generalizability of our method, we conduct evaluation on the MipNeRF dataset. As this dataset does not contain ground truths, we thus provide only visual comparison. As shown in Figure~\ref{fig:mipnerf_effect}, our method exhibits clear advantage over the compared methods in handling complex real-world lighting and material interactions.

\vspace{0.5em}
\noindent{\noindent\textbf{Evaluation on the DL3DV and FIPT datasets.}
Besides MipNeRF, we also evaluate our method on two additional real-world datasets, \ie, DL3DV and FIPT, which contain complex real-world environments with large-area lights and extended luminaries. As shown in Table~\ref{table:realscene}, our method reports better quantitative results on the two datasets, manifesting its effectiveness.

\subsection{More Analysis}

To examine the contribution of each component in our method, we conduct ablation studies on the following modules: 
(i) the geometry and semantic reconstruction, 
(ii) the hybrid illumination model and the deshadowing module, and 
(iii) the material constraints based on view and illumination invariance. In addition, we evaluate the expressiveness of our lighting model and the robustness of our method to unreliable pretrained model priors.

\vspace{0.5em}
\noindent\textbf{Effect of geometry and semantic reconstruction.}
To verify the effectiveness of our initialization strategy, we separately analyze the contribution of the normal prior and the semantic prior. Quantitative results are reported in Table~\ref{table:ablation}, where we can see that both priors are beneficial to the success of our method. Please refer to the supplementary material for visual demonstrations.

\vspace{0.5em}
\noindent\textbf{Effect of hybrid illumination model and deshadowing module.}
We in Table~\ref{table:ablation} and Figure~\ref{fig:ablation_albedo_estimation} validate the effectiveness of our hybrid illumination model and the deshadowing module. As shown, the SGM point lights enables our model to better capture high-frequency illumination effects, while the deshadowing module helps effectively remove cast shadows from the estimated materials, leading to cleaner albedo prediction.

\vspace{0.5em}
\noindent\textbf{Effect of material-invariant constraints} . Table~\ref{table:ablation} and Figure~\ref{fig:ablation_albedo_estimation} present both quantitative and qualitative results to demonstrate the effectiveness of the material-invariant constraints ($\mathcal{L}_\text{relight}$ and $\mathcal{L}_\text{cross\_semantic}$). As shown, these constraints allow to improve the fidelity of the recovered albedo, and help produce predictions closer to the ground-truth albedos.

\input{tables/realworld_comparison}

\input{images/Expressiveness_SGM}

\vspace{0.5em}
\noindent\noindent\textbf{Robustness to noisy pre-trained model priors.} We also evaluate the robustness of our method to noisy priors from pre-trained model as well as inaccurate camera parameters on the Interiorverse dataset. Table~\ref{table:robustness_analysis} summarizes the numerical results. As shown, introducing random positional perturbations to the Gaussian geometric field or noise to semantic labels and normal priors results in only marginal performance degradation, demonstrating the robustness of our framework. Moreover, our approach also has some tolerance to inaccurate camera poses. By refining Gaussian positions during the initial training stage, our method compensates for suboptimal initializations from sparse-view VGGT and achieves performance on par with the 200-view ``gold standard'' (referred to as GT Cameras).

\begin{figure}[t]
\centering
\includegraphics[width=1.0\columnwidth]{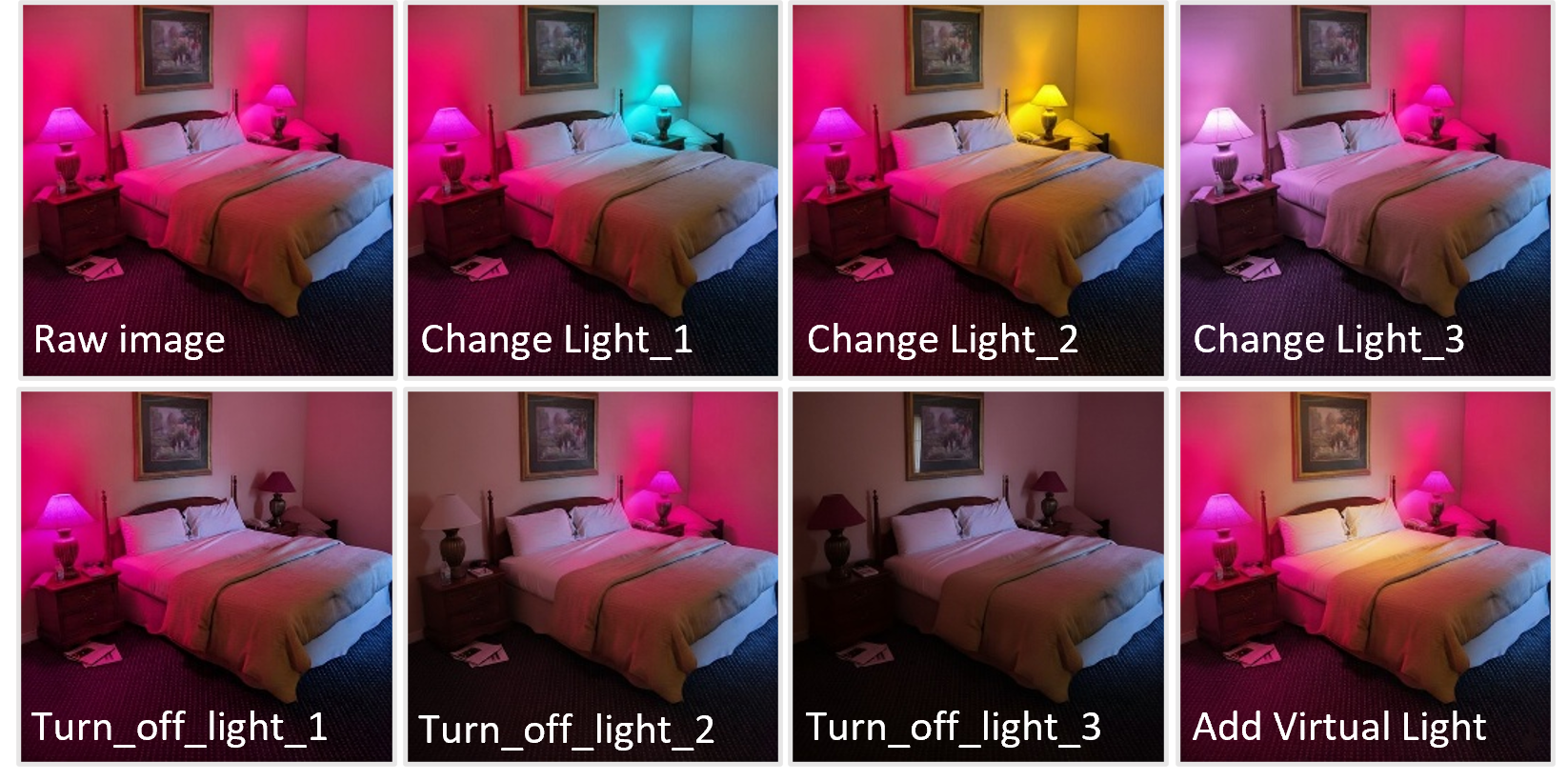} \\
\vspace{-2mm}
\caption{\textbf{Application to lighting control.}}
\label{fig:application}
\end{figure}

\input{tables/robust_analysis}

\input{tables/ablation}

\vspace{0.5em}
\noindent \textbf{Expressiveness of our lighting model.} Figure~\ref{fig:complex_emitter} shows that our SGM-based interior lighting model can effectively model complex indoor emitters, including large area lights and extended luminaries by increasing the number of the initial point lights. 


\vspace{0.5em}
\noindent \textbf{Applications.} As demonstrated in Figure~\ref{fig:application}, since our method allows to recover scene geometry, material properties, and illumination, it enables complex lighting edits, including fine-grained control over light color, individual source toggling, and the seamless integration of virtual lights to modify global lighting. Please see the supplementary material for more results.

%% file: images/mipnerf_comparison.tex
\newcommand{\imgwithlabel}[3]{%
  \begin{overpic}[width=#1]{#2}
    \put(96,4){
      \makebox[0pt][r]{
        \textcolor{white}{\footnotesize\textsf{#3}}%
      }%
    }%
  \end{overpic}%
}

\begin{figure*}[t]
\setlength\tabcolsep{1.25pt}
\centering
\includegraphics[width=\textwidth]{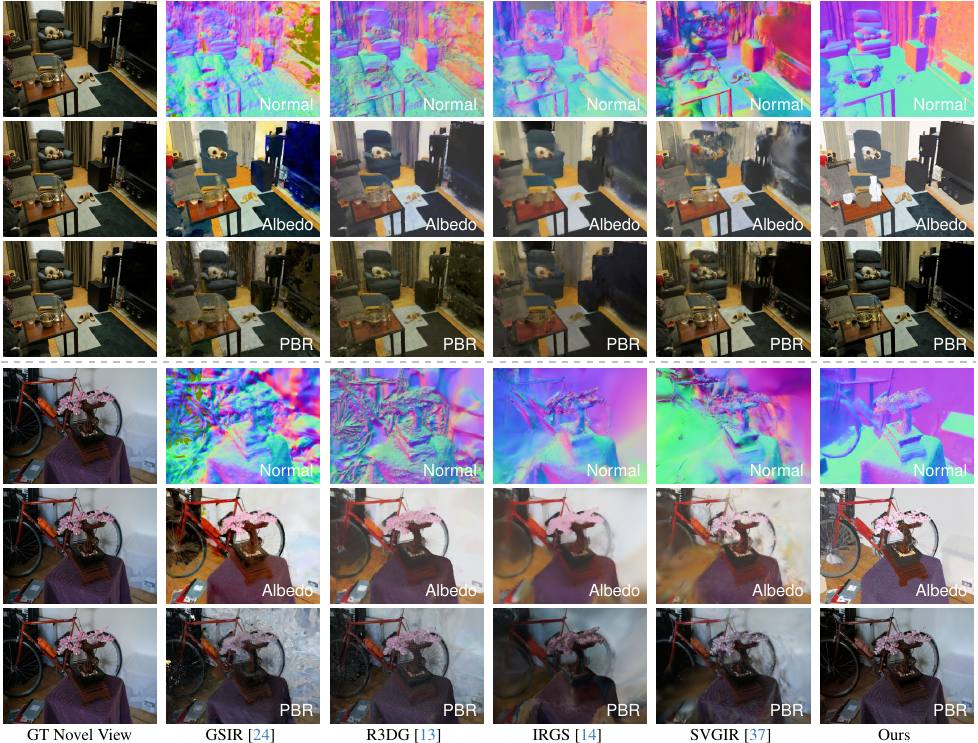}\\
\vspace{-2mm}
\caption{\textbf{Qualitative comparison of inverse rendering on the MipNeRF dataset~\cite{barron2022mip}}. }
\label{fig:mipnerf_effect}
\end{figure*}

%% file: images/ablation_analysis.tex
\begin{figure*}[t]
\setlength\tabcolsep{1.25pt}
\centering
\includegraphics[width=\textwidth]{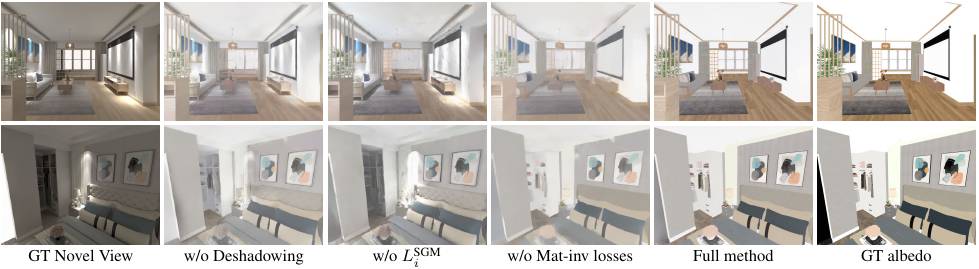}\\
\vspace{-3mm}
\caption{\textbf{Qualitative ablation study on the Interiorverse dataset.} Note, ``Mat-inv losses'' refer to $\mathcal{L}_\text{relight}$ and $\mathcal{L}_\text{cross\_semantic}$.}
\label{fig:ablation_albedo_estimation}
\end{figure*}

%% file: tables/realworld_comparison.tex
\begin{table*}[!t]
\centering
\begin{minipage}{0.93\textwidth} 
    \centering
    \caption{\textbf{Quantitative comparison with previous methods in terms of novel view synthesis on real-world datasets.} } 
    \label{table:realscene}
    \vspace{0.0em} 
    
    \resizebox{\textwidth}{!}{ 
        \begin{tabular}{l| ccc | ccc | ccc}
        \hline
        \multirow{2}{*}{Method} & \multicolumn{3}{c|}{DL3DV} & \multicolumn{3}{c|}{FIPT} & \multicolumn{3}{c}{MipNeRF}\\
        & PSNR$\uparrow$ & SSIM$\uparrow$ & LPIPS$\downarrow$ & PSNR$\uparrow$ & SSIM$\uparrow$ & LPIPS$\downarrow$ & PSNR$\uparrow$ & SSIM$\uparrow$ & LPIPS$\downarrow$\\
        \hline
        GSIR~\cite{liang2024gs}      & 18.65 & 0.38 & 0.48 & 19.12 & 0.42 & 0.42 & 18.95 & 0.40 & 0.45 \\
        R3DG~\cite{gao2024relightable}    & 18.22 & 0.35 & 0.52 & 18.85 & 0.39 & 0.47 & 18.61 & 0.37 & 0.49 \\
        SVGIR~\cite{sun2025svg}   & 17.96 & 0.31 & 0.59 & 18.53 & 0.36 & 0.54 & 18.42 & 0.35 & 0.56 \\
        GeoSplat~\cite{ye2025geosplatting}& 18.98 & 0.40 & 0.44 & 19.75 & 0.45 & 0.38 & 19.44 & 0.43 & 0.41 \\
        IRGS~\cite{gu2025irgs}    & 18.34 & 0.36 & 0.51 & 18.96 & 0.40 & 0.45 & 18.72 & 0.38 & 0.48 \\
        \hline
        Ours         & \textbf{19.31} & \textbf{0.42} & \textbf{0.41} & \textbf{20.10} & \textbf{0.48} & \textbf{0.35} & \textbf{19.73} & \textbf{0.46} & \textbf{0.38} \\
        \hline
        \end{tabular}
    }
\end{minipage}
\end{table*}

%% file: images/Expressiveness_SGM.tex
\begin{figure*}[htbp]
  \centering
  \begin{subfigure}{0.48\linewidth}
    \centering
    \includegraphics[width=\linewidth]{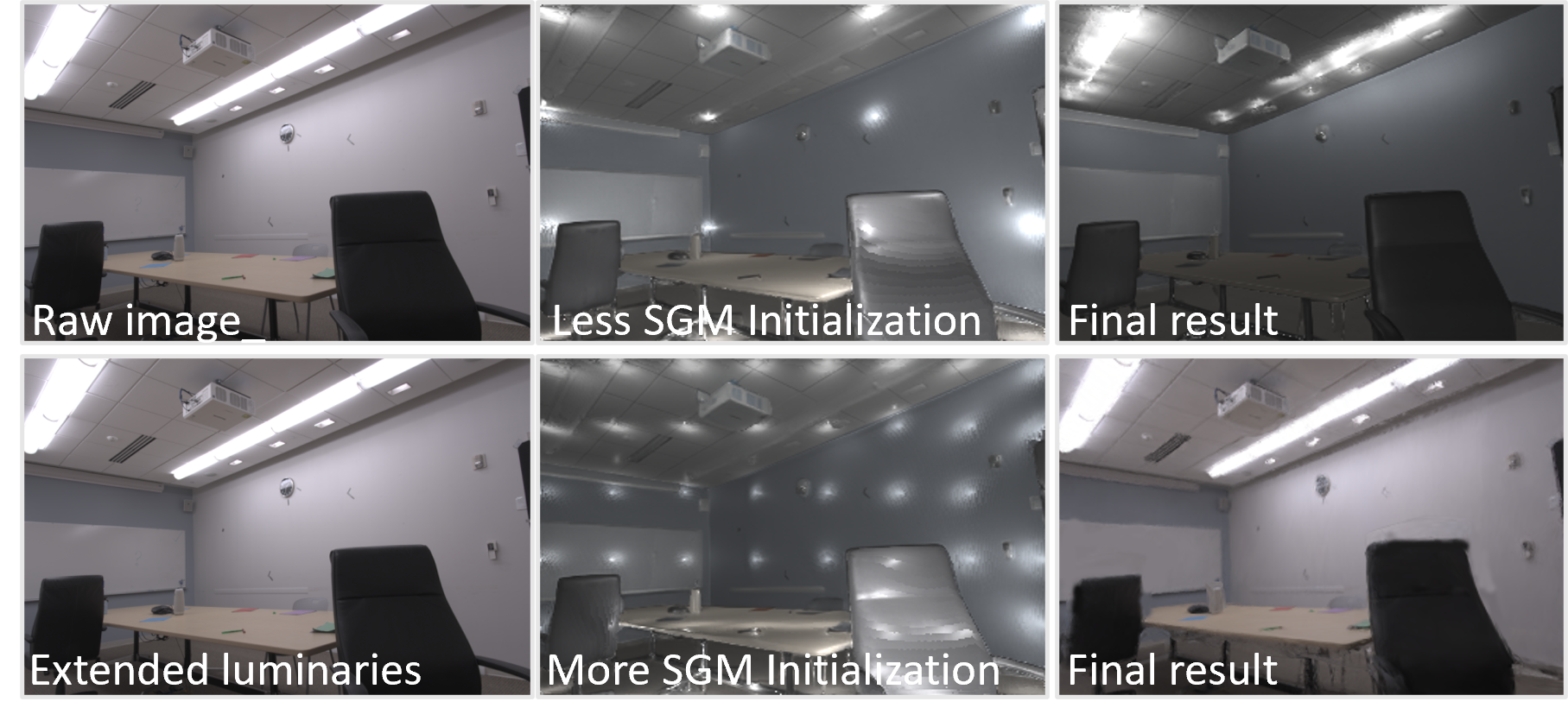}
    \vspace{-6mm} 
    \label{fig:light_source}
  \end{subfigure}%
  \hspace{1.5mm}
  \begin{tikzpicture}
    \draw[dash pattern=on 3pt off 3pt, gray!50, thick] (0,0) -- (0,3.5cm); 
  \end{tikzpicture}
  \hspace{0.1mm}
  \begin{subfigure}{0.48\linewidth}
    \centering
    \includegraphics[width=\linewidth]{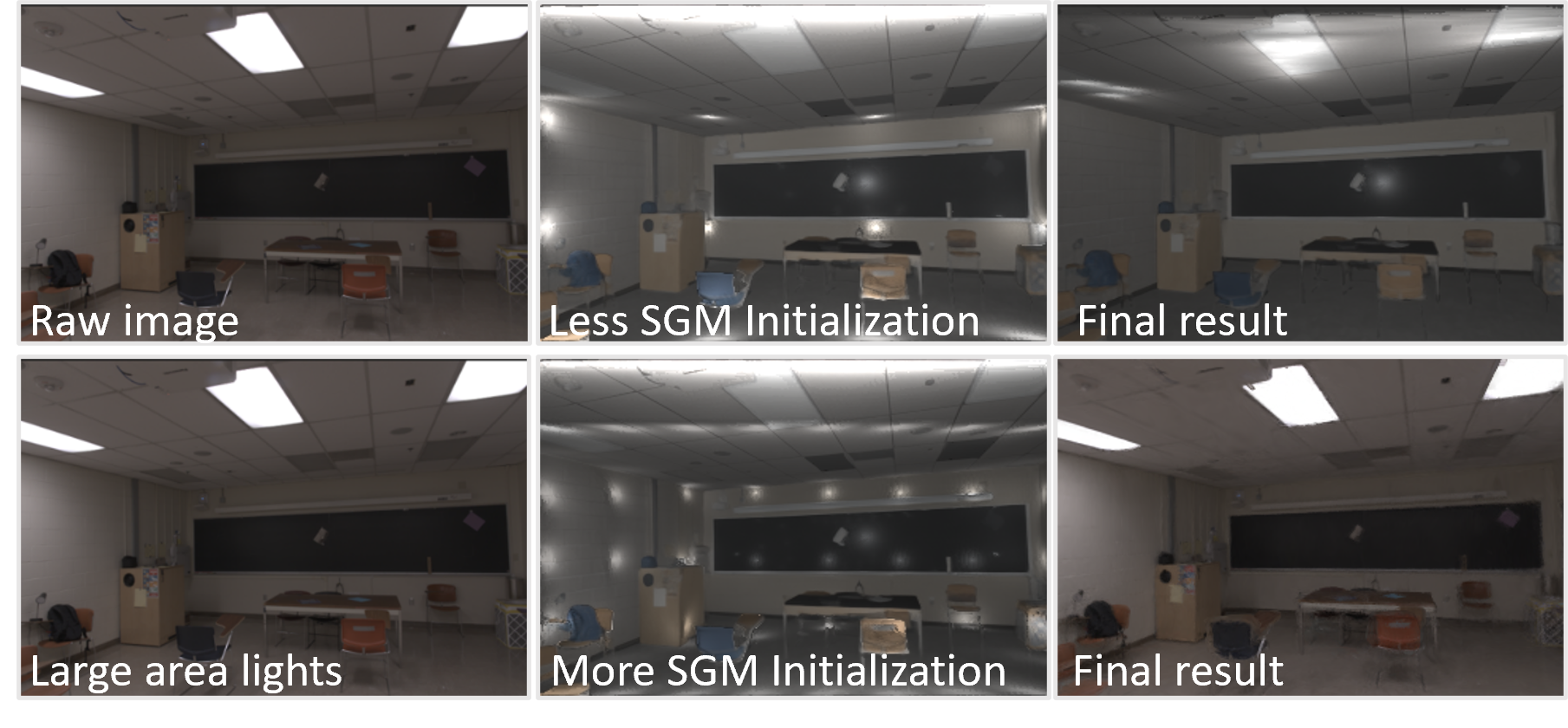}
    \vspace{-6.1mm}
    \label{fig:scene_composition}
  \end{subfigure}
  \vspace{-1mm}
  \caption{\textbf{Expressiveness of our SGM-based interior lighting model for complex emitters.}}
  \label{fig:complex_emitter}
\end{figure*}

%% file: tables/robust_analysis.tex
\begin{table}[!t]
\centering
\vspace{-0.5em}
\caption{
\textbf{Robustness analysis on Interiorverse dataset.}  ``w/ Geom noise'' denotes introducing random positional perturbations to the Gaussian geometric field, while ``w/ normal noise'' and ``w/ semantic noise'' mean adding random noise to normal priors and semantic labels. ``GT Cameras'' refers to camera poses reconstructed from 200 views via VGGT.}
\vspace{-0.5em}
\resizebox{1.0\columnwidth}{!}{
\begin{tabular}{l c c c c c c} 
\hline
\multirow{2}{*}{Method} & \multicolumn{3}{c}{Albedo} & \multicolumn{3}{c}{NVS For PBR} \\ 
& PSNR$\uparrow$ & SSIM$\uparrow$ & LPIPS$\downarrow$ & PSNR$\uparrow$ & SSIM$\uparrow$ & LPIPS$\downarrow$ \\ 
\hline
w/ Geom noise   & 17.5 & 0.73 & 0.31 & 21.6 & 0.75 & 0.45 \\
w/ normal noise   & 17.5 & 0.73 & 0.31 & 21.4 & 0.74 & 0.46 \\
w/ semantic noise   & 17.6 & 0.75 & 0.30 & 21.7 & 0.75 & 0.45 \\
w/ GT Cameras   & 17.7 & 0.75 & 0.28 & 21.7 & 0.76 & 0.42 \\
Ours & \textbf{17.7} & \textbf{0.76} & \textbf{0.29} & \textbf{21.7} & \textbf{0.76} & \textbf{0.43} \\
\hline
\end{tabular}
}
\label{table:robustness_analysis}
\end{table}

%% file: tables/ablation.tex
\begin{table}[!t]
\centering
\caption{
\textbf{Ablation study on the Interiorverse dataset.} ``Deshadowing'' refers to the deshadowing module.
\vspace{-0.5em}
}
\resizebox{1.00\columnwidth}{!}{
\begin{tabular}{l|c c c|c c c}
\hline
\multirow{2}{*}{Method} &
\multicolumn{3}{c|}{Albedo} &
\multicolumn{3}{c}{NVS For PBR}  \\
&PSNR$\uparrow$ & SSIM$\uparrow$ & LPIPS$\downarrow$ &
PSNR$\uparrow$ & SSIM$\uparrow$ & LPIPS$\downarrow$ \\
\hline
w/o $\mathcal{L}_{normal}$ &
16.90&0.73&0.32 &
20.35 & 0.70 & 0.32 \\
w/o $\mathcal{L}_{corresponding}$ &
16.80 & 0.71&0.32 &
19.78 & 0.70  & 0.30 \\
w/o $L_i^\text{SGM}$ &
14.78&0.65&0.40& 
20.03 & 0.69 & 0.37 \\
w/o Deshadowing &
16.22&0.68&0.35&
20.71 & 0.72 & 0.35 \\
w/o $\mathcal{L}_{relight}$  &
17.20 &0.73&0.34& 
20.30 & 0.73 & 0.34 \\
w/o $\mathcal{L}_{cross\_semantic}$  &
17.10 & 0.71 & 0.35 &
20.20 & 0.71 & 0.32 \\
\hline
Ours  &
\textbf{17.70} & \textbf{0.76} & \textbf{0.29} & 
\textbf{20.90}&\textbf{0.73}&\textbf{0.31}\\
\hline
\end{tabular}}

\label{table:ablation}
\end{table}

%% file: sec/5_conclusion.tex
\section{Conclusion}

In this work, we propose SGS-Intrinsic, a method for indoor inverse rendering of sparse-view images. The core idea lies in constructing a robust geometric field and encouraging illumination and material representations to remain as disentangled as possible within the optimization space. To achieve this, we develop an effective two-stage training strategy. Extensive experiments demonstrate the effectiveness of our approach.

%% file: sec/X_supple.tex
\setcounter{section}{0}

\maketitlesupplementary
This supplementary material provides a detailed description of our method's implementation and more 
visual comparison results.

\section{Representation}
In the vanilla 3DGS, each 3D Gaussian utilizes learnable parameters $\bm{T} = \{\bm{p}, \bm{s}, \bm{q} \}$ and $\bm{C} = \{\alpha, \bm{f}_{c}\}$ to describe its geometric properties and volumetric appearance, respectively.
Here, $\bm{p}$ denotes the position vector, $\bm{s}$ denotes the scaling vector, $\bm{q}$ denotes the unit quaternion for rotation, $\alpha$ denotes the opacity, and $\bm{f}{c}$ denotes the spherical harmonics (SH) coefficients for view-dependent color.
In SGS-Intrinsic, we extend $\bm{C}$ to $\{\alpha, \bm{f}_{c}, \bm{f}_{s}, \bm{n}, \bm{a}, \bm{r}\}$ to describe the material and semantic properties of the 3D Gaussian, where $\bm{a}$ and $\bm{r}$ denote the albedo and roughness values, $\bm{n}$ denotes the Normal, and $\bm{f}_{s}$ denotes the semantic feature.

\section{Implementation Details and Constraints of Hybrid Illumination Models}
We employ physically-based rendering (PBR) to model the view-dependent appearance. Specifically, we utilize the CookTorrance model to formulate the bidirectional reflectance distribution function (BRDF) fr:
\begin{equation}
\label{eq:brdf:supp}
\begin{aligned}
f_r(\omega_i, \omega_o) &= 
\underbrace{
(1 - M) \frac{A}{\pi}
}_{\text{diffuse component}} +
\underbrace{
\frac{DFG}{4(n \cdot \omega_i)(n \cdot \omega_o)}
}_{\text{specular component}}, \\
h &= \text{normalize}(\omega_o + \omega_i),\\
F_0 &= (1 - M) \cdot 0.04 + M \cdot A, \\
D(n, h) &= \frac{R^4}{\pi \left(
n \cdot h \ (R^4 - 1) + 1
\right)^2},\\
F(\omega_i, n) &= F_0 + (1 - F_0) \left(
1 - n \cdot \omega_i
\right)^5,\\
G(\omega_o, \omega_i, h) &= G_1(\omega_o, h) \cdot G_1(\omega_i, h),\\
G_1(n, h) &= \frac{1}{1 + n \cdot h \sqrt{R^4 + n \cdot h - R^4 \cdot n \cdot h}},
\end{aligned}
\end{equation}
where $A$, $R$, and $M$ denote the albedo, roughness, and metallicity, respectively. The normal distribution function (NDF) $D$, fresnel function $F$, and geometry function $G$ are derived from physical materials.

\begin{algorithm}[t]
  \caption{ Pseudo view sampling algorithm}
  \label{alg:view_spline}
  \begin{algorithmic}[1] 
    \Require Training views $V_{\text{train}}$, test views $V_{\text{test}}$;
             interpolation counts $N_{\text{train}},N_{\text{test}}$;
             distance weights $\lambda_t,\lambda_r$
    \Statex
    \Function{PoseDist}{\ensuremath{(q_1,t_1),(q_2,t_2)}}
      \State $D \gets \lambda_t \|t_1-t_2\|_2 + \lambda_r \arccos\big(2 (q_1^\top q_2)^2 - 1\big)$
      \State \Return $D$
    \EndFunction
    
    \Statex
    \For{each input view $(q,t)$ in $V_{\text{train}}$}
      \State Find nearest training view $(q^{\text{nn}}_{\text{tr}},t^{\text{nn}}_{\text{tr}})$ in $V_{\text{train}}$ using \Call{PoseDist}{}
      \State Generate $N_{\text{train}}$ spline-interpolated pseudo-views
      \Statex\hskip\algorithmicindent $\mathcal{S}_{\text{train}} \gets \text{CubicSpline}\big((q,t),(q^{\text{nn}}_{\text{tr}},t^{\text{nn}}_{\text{tr}})\big)$
      \State Find nearest test view $(q^{\text{nn}}_{\text{te}},t^{\text{nn}}_{\text{te}})$ in $V_{\text{test}}$ using \Call{PoseDist}{}
      \State Generate $N_{\text{test}}$ spline-interpolated pseudo-views
      \Statex\hskip\algorithmicindent $\mathcal{S}_{\text{test}} \gets \text{CubicSpline}\big((q,t),(q^{\text{nn}}_{\text{te}},t^{\text{nn}}_{\text{te}})\big)$
      \State Randomly sample $v_{\text{tr}}\sim\mathcal{S}_{\text{train}},\; v_{\text{te}}\sim\mathcal{S}_{\text{test}}$
      \State Construct view-sequence $\mathcal{V} \gets \{v_{\text{tr}},(q,t),v_{\text{te}}\}$
      \State Feed sequence $\mathcal{V}$ into the SAM2 model
    \EndFor
  \end{algorithmic}
\end{algorithm}

We employ a mixture model of a set of spherical Gaussians (SGMs) to represent the localized highlight illumination as $L^\text{SGMs}_{o}(\bm{x}, \bm{\omega}_o)$:
\begin{equation}
\begin{aligned}
L^\text{SGMs}_{o}(\bm{x}, \bm{\omega}_o) &= \sum_{i}^{N_\text{light}} \frac{{f_r}^\text{SGM}_{i} (\bm{n} \cdot \bm{\omega}_{i})*\text{V}_i}{d_i^2} \sum_{j}^{N_\text{sg}} W_{i, j} SG(j),
\end{aligned}
\end{equation}
where $d_i$ denotes the distance from the $i$-th spherical Gaussian mixture to the surface point $\bm{x}$. ${f_r}^\text{SGM}_{i}$ denotes the BRDF function, and $SG$ denotes the spherical Gaussian Function, respectively. Each spherical Gaussian mixture contains $N_{\text{sg}}$ spherical Gaussians.

Following ~\cite{liang2024gs}, we represent environment lighting as a learnable cubemap with size $6 \times 512 \times 512 \times 3$. Using split-sum approximation, the environment radiance can be decomposed as:
\begin{equation}
\small
\begin{gathered}
\label{eq:env_diff_spec_approx}
L^\text{env}_{o}(\bm{x}, \bm{\omega}_o) =
L^\text{env}_{o\text{-diff}} + L^\text{env}_{o\text{-spec}},\\
L^\text{env}_{o\text{-diff}} \approx K^\text{env}_\text{diff} I^\text{env}_\text{diff},\quad
K^\text{env}_\text{diff} = (1 - M) \frac{A}{\pi},\\
\begin{aligned}
L^\text{env}_{o\text{-spec}}
\approx& \underbrace{
\int_{\Omega} \frac{DFG}{4 (\bm{n} \cdot \bm{\omega}_o)} \, \text{d}\bm{l}
}_{
\text{Environment BRDF }(K^\text{env}_\text{spec})
}
\cdot\underbrace{
\int_{\Omega} 
DL_i(\bm{l}) (\bm{l} \cdot \bm{n}) \, \text{d}\bm{l}.
}_{
\text{Pre-Fil. Env. Map }(I^\text{env}_\text{spec})
}
\end{aligned}
\end{gathered}
\vspace{-0.1cm}
\end{equation}

\input{images/tensoir_compare}

We further introduce two regularization terms to encourage compact and stable light representations:
\begin{align}
    L_{\text{pos}} &= \sum_{i}^{N_{\text{light}}} \frac{1}{d_{i,\text{near}}}, \\
    L_{\text{val}} &= \sum_{i}^{N_{\text{light}}} \sum_{j}^{N_{\text{sg}}} w_{i,j}.
\end{align}

\section{Pseudo Views Sampling Algorithm}
Please refer to the pseudocode in Algorithm \ref{alg:view_spline} for further details.

\section{Deshaowing Module design}
The deshadowing module incorporates a hash encoder to perform multi-resolution encoding of the 3D Gaussian field, along with a lightweight MLP decoder that predicts the occlusion relationship of a given spatial point relative to a point light source. The specific details of this module are presented in Table \ref{table:deshadowing}.

\begin{table}[h]
\centering
\caption{Architecture of Deshadowing Module}
\label{table:deshadowing}
\vspace{-0.2cm}
\begin{tabular}{ccc}
\hline
\textbf{Module} & \textbf{Parameter} & \textbf{Value} \\
\hline
\multirow{5}{*}{HashEnc} & Number of levels & 16 \\
 & Max. entries per level  & $2^{19}$ \\
 & feature dimensions & 2 \\
 & Coarsest resolution & 32 \\
 & Finest resolution & 4096 \\
\hline
\multirow{3}{*}{MLP} & MLP layers & $36 \times 32 \times 32 \times 1$ \\
 & Initialization & Kaiming-uniform \\
 & Final activation & Sigmoid \\
\hline
\end{tabular}
\end{table}

\section{Training Details}

We implement our method and conduct all experiments on a single NVIDIA RTX 4090 GPU.  

For the first-stage geometric reconstruction, we set the total number of training iterations to 7000.  
Virtual view sampling and the semantic consistency constraint between virtual and real views are activated after 2000 iterations.  
The learnable semantic feature dimension attached to each Gaussian is set to 32.  
We employ CLIP ViT-B/16~\cite{radford2021learning} for semantic feature extraction and SAM2 Hiera-L ~\cite{ravi2024sam} for region mask generation.

For the second stage, corresponding to the inverse rendering phase, the number of training iterations is set to 3000.  
At the beginning of this stage, we initialize the point light sources represented by Spherical Gaussian Mixtures (SGMs) uniformly within the scene’s axis-aligned bounding box (AABB) of range $[-1.5, 1.5]$.  
To balance representational expressiveness and computational cost, we place $[4,4,4]$ point lights in total, resulting in 64 point light sources.  
Each point light source contains $n_{SG}=12$ spherical Gaussians.  
The illumination invariance constraint is introduced after 8000 iterations, and both the self-view invariance and multi-view invariance constraints are applied every three iterations.

\begin{table}[!t]
\centering

\caption{\textbf{Quantitative comparison on the TensoIR dataset.}}
\vspace{-0.2cm}

\resizebox{0.99\columnwidth}{!}{
\begin{tabular}{@{}l|ccc|ccc}
\toprule
\multirow{2}{*}{Method} & \multicolumn{3}{c@{}|}{Albedo} & \multicolumn{3}{c@{}}{NVS For PBR} \\ 
& PSNR$\uparrow$ & SSIM$\uparrow$ & LIPPS$\downarrow$ &
PSNR$\uparrow$ & SSIM$\uparrow$ & LIPPS$\downarrow$ 
\\
\hline
GSIR~\cite{liang2024gs}&
23.67&	0.90&	0.11 & 23.98&0.82&0.14 
\\
R3DG~\cite{gao2024relightable}& 22.46& 0.89&	0.12 & 25.11&	0.85 &	0.12
\\
SVGIR~\cite{sun2025svg}& 24.98& 0.90&	0.13 & 21.05&	0.68&	0.19
\\
GeoSplat~\cite{ye2025geosplatting}& 24.31& 0.89&	0.13 & 25.45&	\textbf{0.87}&	0.12
\\
IRGS~\cite{gu2025irgs}& 22.57& 0.86&	0.14 & 25.32&	0.85&	0.13
\\
Ours  &
\textbf{25.65} & \textbf{0.93} & \textbf{0.10}  &
\textbf{26.02} & \textbf{0.87} & \textbf{0.11} 
\\
\bottomrule
\end{tabular}
}
\label{table:tensoir}
\end{table}


\section{Evaluation on the TensoIR dataset}
We also evaluate our approach on the TensoIR dataset. The quantitative and qualitative results are presented in Table~\ref{table:tensoir} and Figure~\ref{fig:tensoir_compare}, respectively. As shown, our method achieves leading performance on object-level data, further validating its robustness and generalization capability.

\begin{table}[!t]
  \centering
  \vspace{-1.8mm}
  \caption{Training time breakdown} 
  \vspace{-0.3cm}
  \resizebox{\columnwidth}{!}{
    \begin{tabular}{lccccc} 
      \toprule
      & VGGT & LSeg & StableNorm & RGBX&Training \\
      \midrule
      Time & $\sim$20s & $\sim$2min & $\sim$10s & $\sim$30s&$\sim$40min \\
      \bottomrule
    \end{tabular}
  }
  \label{table:process_alltime}
\end{table}

\section{Qualitative Results of Geometry and Semantic Reconstruction}
\begin{figure}[h]
\centering
\includegraphics[width=1.0\columnwidth]{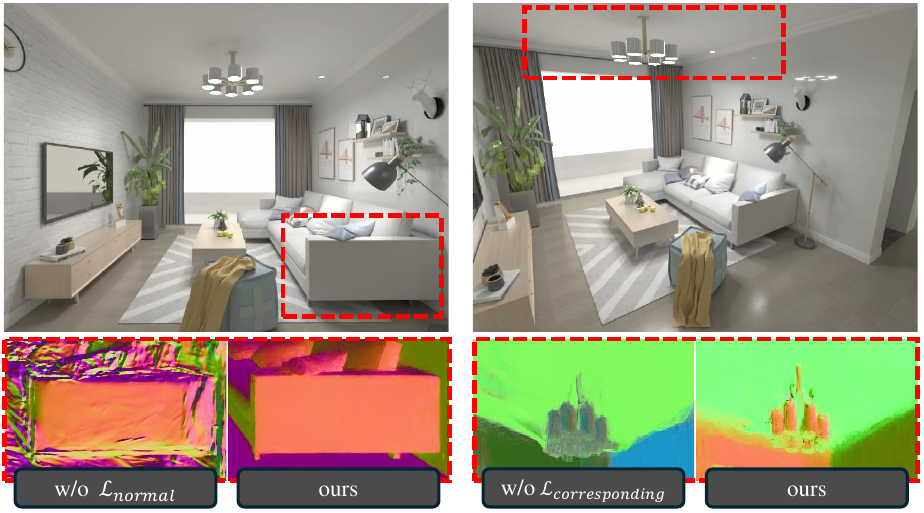} \\
\caption{\textbf{Effect of normal and semantic priors.}}
\label{fig:ablation}
 \vspace{-2mm}
\end{figure}

Figure~\ref{fig:ablation} demonstrates the respective impacts of the normal prior and the semantic prior on the reconstruction results. The normal priors provide smoother geometric surfaces, significantly improving the quality of geometry reconstruction. The semantic priors, together with multi-view semantic consistency supervision, effectively alleviate the geometric reconstruction issues of 3D Gaussians in textureless regions.

\section{Breakdown of the total time} 
The total duration of the preprocessing workflow is summarized in Table ~\ref{table:process_alltime}, amounting to a cumulative processing time of about 3 minutes.


\section{Visualizing multi-resolution features}
\begin{figure}[h]
  \centering
  \includegraphics[width=1.0\linewidth]{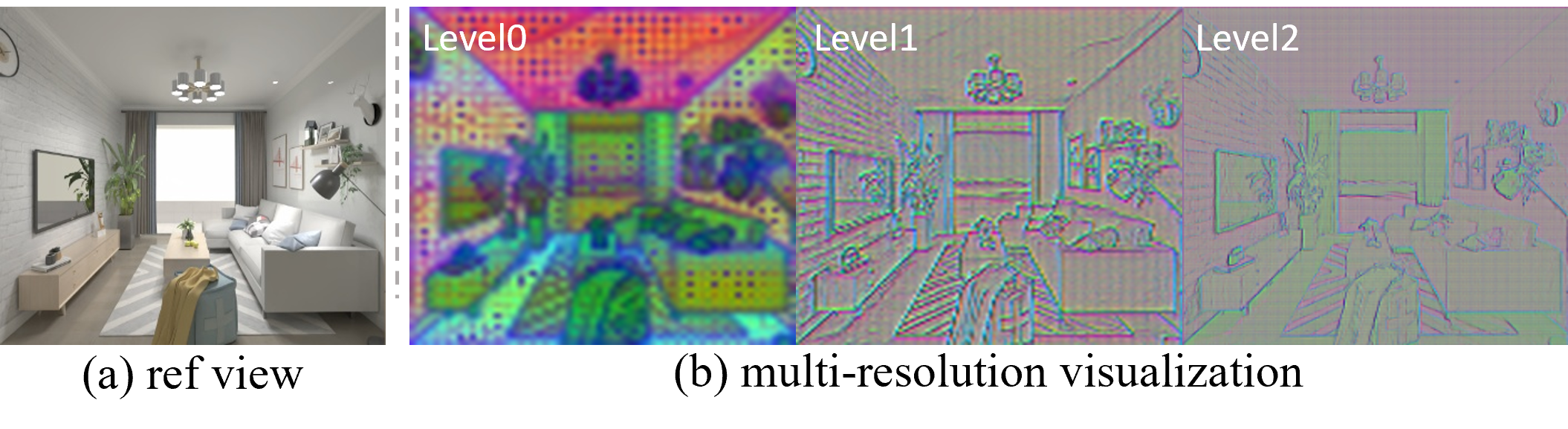} \\
  \vspace{-2mm}
   \caption{\textbf{Multi-resolution feature visualization and comparison with GT view}}
   \label{fig:multihash_visualization}
   \vspace{-5.8mm}
\end{figure}

As shown in Figure~\ref{fig:multihash_visualization}, the multi-resolution features are highly correlated with the scene geometry.

\section{More Visual Comparison}
Figure~\ref{fig:albedo_supple1},~\ref{fig:pbr_supple1},~\ref{fig:normal_supple2} provide more visual comparison on material estimation and Novel View PBR Rendering result. Figure~\ref{fig:illu_decomp} presents the detailed decomposition results of our hybrid illumination model, while  Figure~\ref{fig:application2} provides additional application results.

\input{sec/X_suppl}

%% file: images/tensoir_compare.tex
\begin{figure}[t]
\setlength\tabcolsep{1.25pt}
\centering
\includegraphics[width=\linewidth]{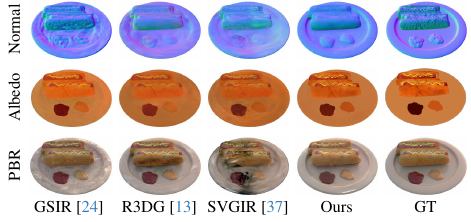}\\
\vspace{-2mm}
\caption{\textbf{Qualitative evaluation on the TensoIR dataset.}}
\label{fig:tensoir_compare}
\end{figure}

%% file: sec/X_suppl.tex
\begin{figure*}[t]
\setlength\tabcolsep{1pt}
\centering
\scriptsize 

\includegraphics[width=\textwidth]{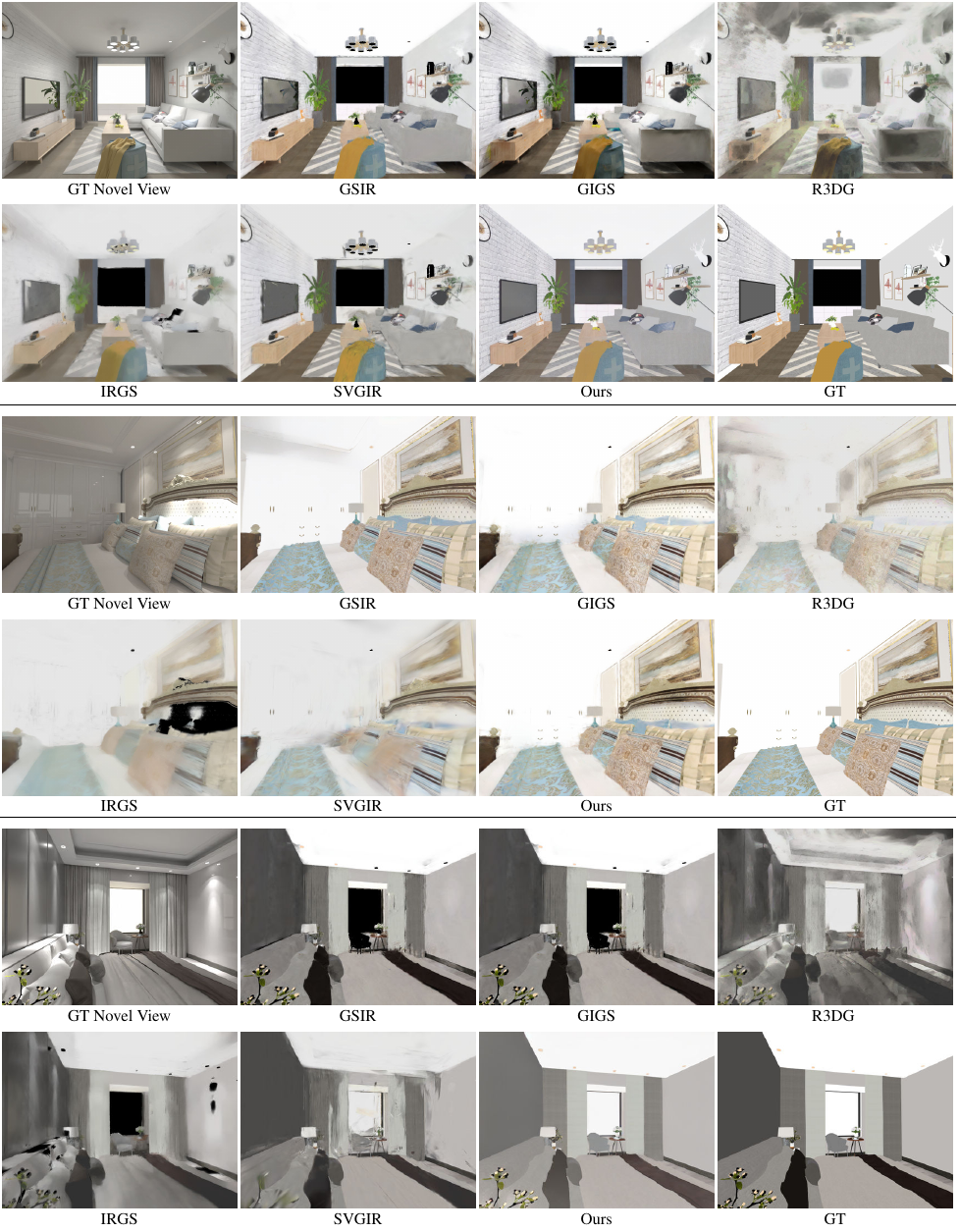}\\
\vspace{-2mm}

\caption{\textbf{More qualitative comparison of albedo estimation on the synthetic InteriorVerse dataset.}}
\label{fig:albedo_supple1}
\end{figure*}

\begin{figure*}[t]
\setlength\tabcolsep{1pt}
\centering
\scriptsize 

\includegraphics[width=\textwidth]{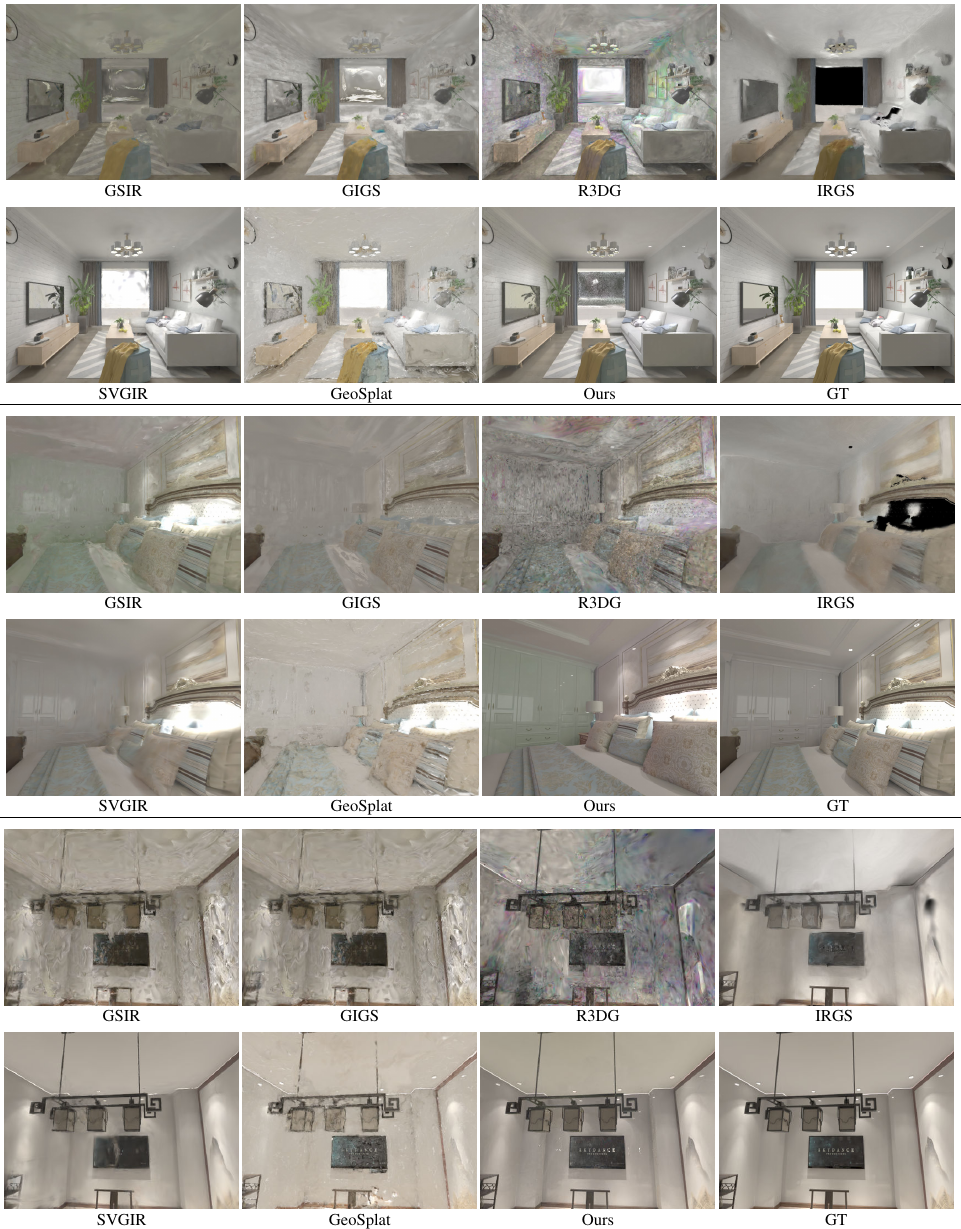}\\
\vspace{-2mm}

\caption{\textbf{More qualitative comparison of PBR Rendered Novel View on the synthetic InteriorVerse dataset.}}
\label{fig:pbr_supple1}
\end{figure*}

\begin{figure*}[t]
\setlength\tabcolsep{1pt}
\centering
\scriptsize 

\includegraphics[width=\textwidth]{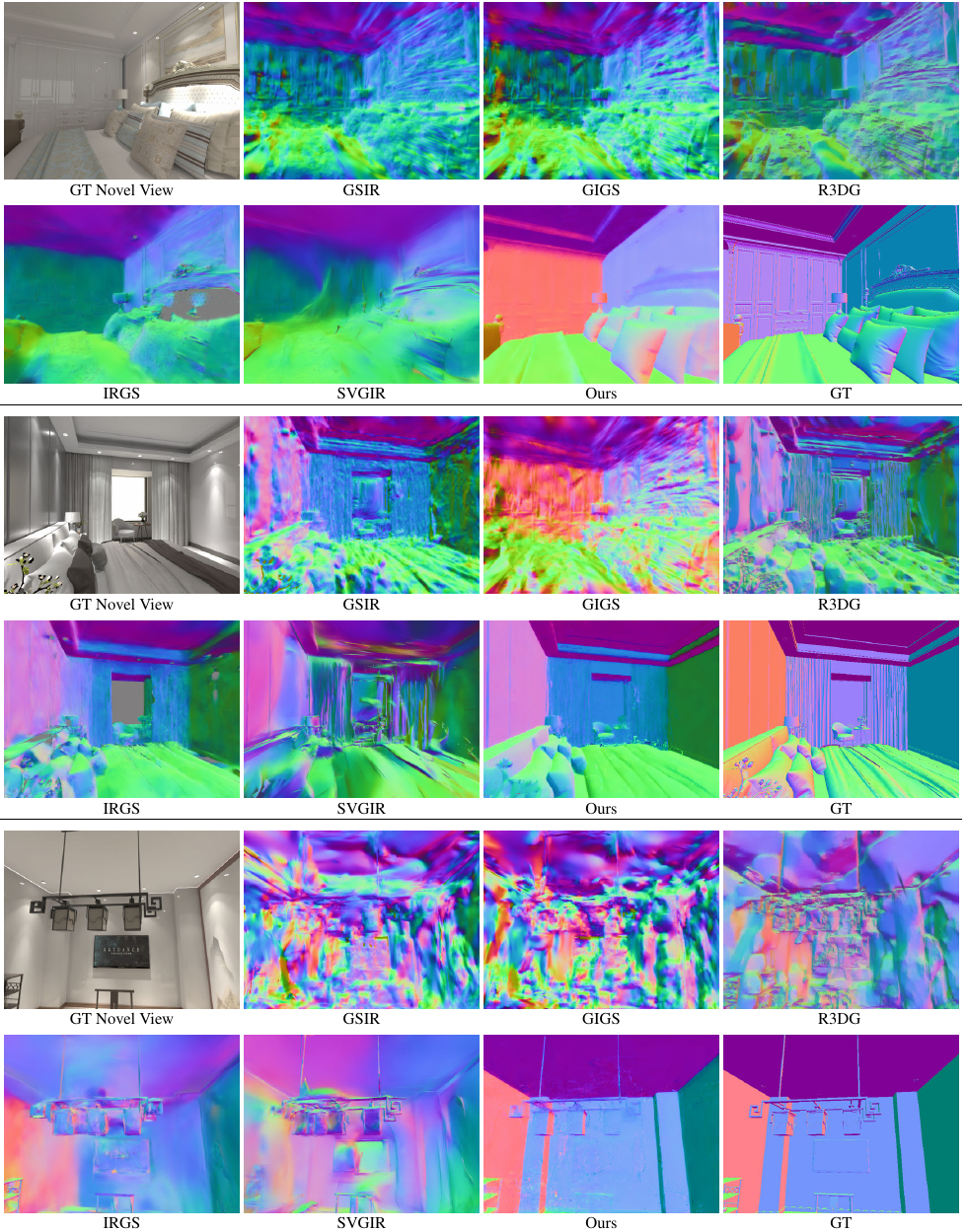}\\
\vspace{-2mm}

\caption{\textbf{More qualitative comparison of albedo estimation on the synthetic InteriorVerse dataset.}}
\label{fig:normal_supple2}
\end{figure*}

\begin{figure*}[t]
\setlength\tabcolsep{1pt}
\centering
\scriptsize 

\includegraphics[width=\textwidth]{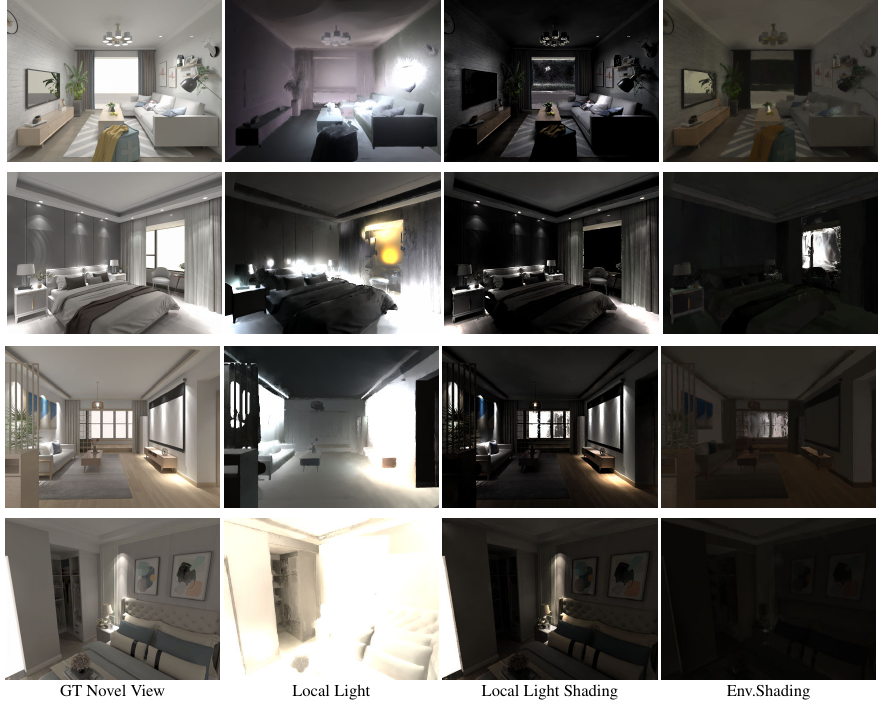}\\
\vspace{-2mm}
\caption{\textbf{More illumination decomposition results on the synthetic InteriorVerse dataset.}}
\label{fig:illu_decomp}
\end{figure*}

\begin{figure*}[t]
\centering
\includegraphics[width=1.0\textwidth]{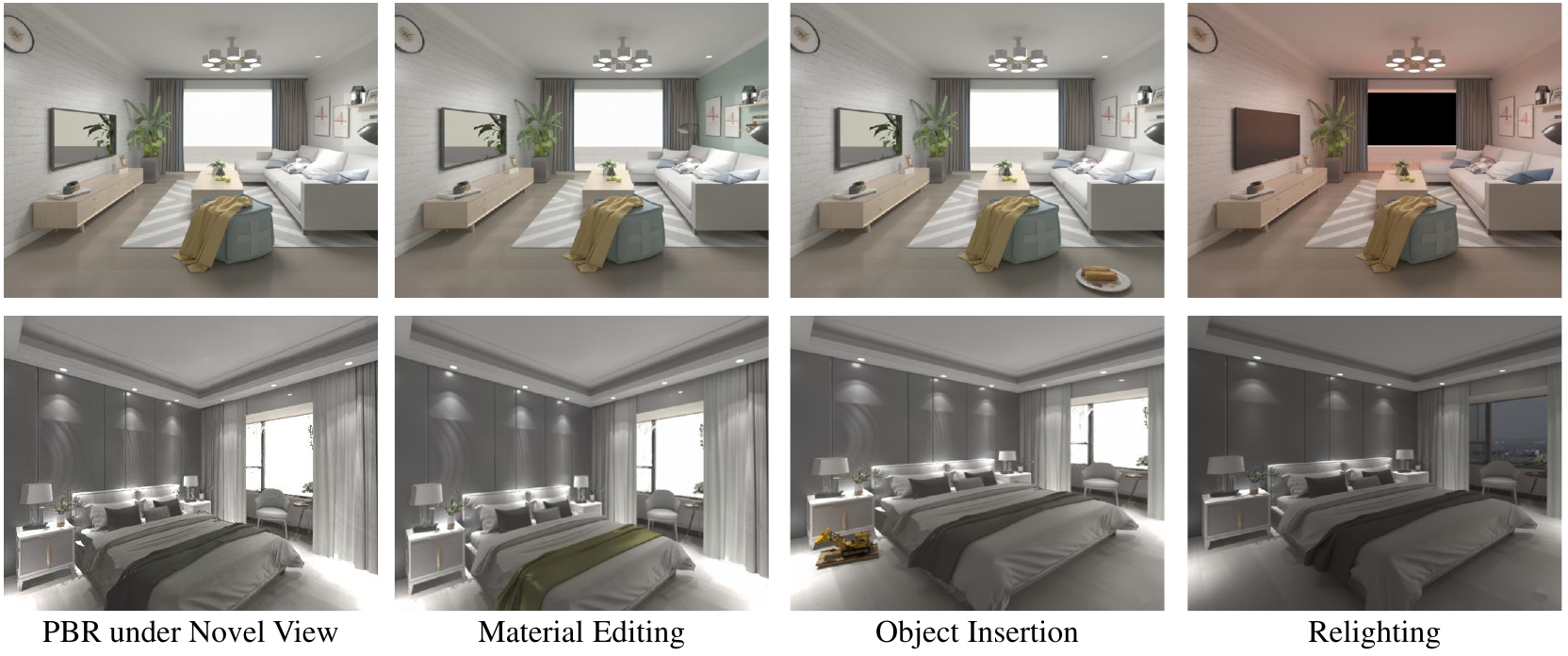} \\
\vspace{-2mm}
\caption{\textbf{More results of object insertion, material and light editing} }
\label{fig:application2}
\end{figure*}